\documentclass{article}

\usepackage{times}
\usepackage{graphicx} 
\usepackage{subfigure} 

\usepackage{natbib}

\usepackage{algorithm}
\usepackage{algorithmic}
\usepackage{multirow}

\usepackage{hyperref}

\usepackage[accepted]{icml2016}

\usepackage{hyperref}
\usepackage{amsmath}
\usepackage{amssymb}
\usepackage[dvipsnames]{xcolor}
\DeclareMathOperator*{\argmax}{arg\,max}

\newcommand{\xx}{{\mathbf{x}}}

\newcommand{\cX}{\mathcal{X}}

\newcommand{\RR}{\mathbb{R}}

\newcommand{\figref}[1]{Figure \ref{#1}}

\graphicspath{{../../../figures/}}

\icmltitlerunning{Evaluation System for a Bayesian Optimization Service}

\begin{document} 

\twocolumn[
\icmltitle{Evaluation System for a Bayesian Optimization Service}

\icmlauthor{Ian Dewancker}{ian@sigopt.com}
\icmlauthor{Michael McCourt}{mike@sigopt.com}
\icmlauthor{Scott Clark}{scott@sigopt.com}
\icmlauthor{Patrick Hayes}{patrick@sigopt.com}
\icmlauthor{Alexandra Johnson}{alexandra@sigopt.com}
\icmlauthor{George Ke}{l2ke@uwaterloo.ca}
\icmladdress{SigOpt, San Francisco, CA 94108 USA}

\icmlkeywords{empirical methods, benchmark systems, bayesian optimization}

\vskip 0.3in
]

\begin{abstract} 

Bayesian optimization is an elegant solution to the hyperparameter optimization problem in machine learning.  
Building a reliable and robust Bayesian optimization service requires careful testing methodology and sound statistical analysis.  In this talk we will outline our development of an evaluation framework to rigorously test and measure the impact of changes to the SigOpt optimization service.  We present an overview of our evaluation system and discuss how this framework empowers our research engineers to confidently and quickly make changes to our core optimization engine

\end{abstract}

\section{Introduction}
\label{submission}

SigOpt offers an optimization service to help customers tune complex systems, simulations and models.  Our optimization engine applies several concepts from Bayesian optimization \cite{bergstra2011algorithms, snoek2012practical, Shahriari2015} and machine learning to optimize customers metrics as quickly as possible.  In particular, we consider problems where the maximum is sought for an expensive function $f : \cX \to \RR$,
\begin{equation*}
\xx_{opt} = \argmax_{\xx \in \cX} f(\xx),
\end{equation*}

within a domain $\cX\subset\RR^d$ which is a bounding box.

Hyperparameter optimization for machine learning models is of particular relevance as the computational costs for evaluating model variations is high, $d$ is typically small, and hyperparameter gradients are typically not available.

SigOpt's core optimization engine is a closed-source fork of the open-source MOE project \cite{Clark2014}.  The SigOpt service supports a succinct set of web API endpoints for optimizing objective functions.     
The evaluation system was built with three high level goals in mind:
\begin{itemize} \setlength\itemsep{.0001em}
	\item Capable of performing end-to-end testing of service
	\item Facilitate comparisons between algorithm versions 
	\item Facilitate comparisons against external baselines   
\end{itemize}

Our evaluation system consists of an extensive benchmark suite of test functions, automated analysis of performance metrics, and visualization tools for test summarization.  The system runs using on-demand cloud infrastructure.

\section{Metrics} 
 
\begin{figure}[H]
	\vspace{-2mm}
	\centering
	\includegraphics[width=\columnwidth]{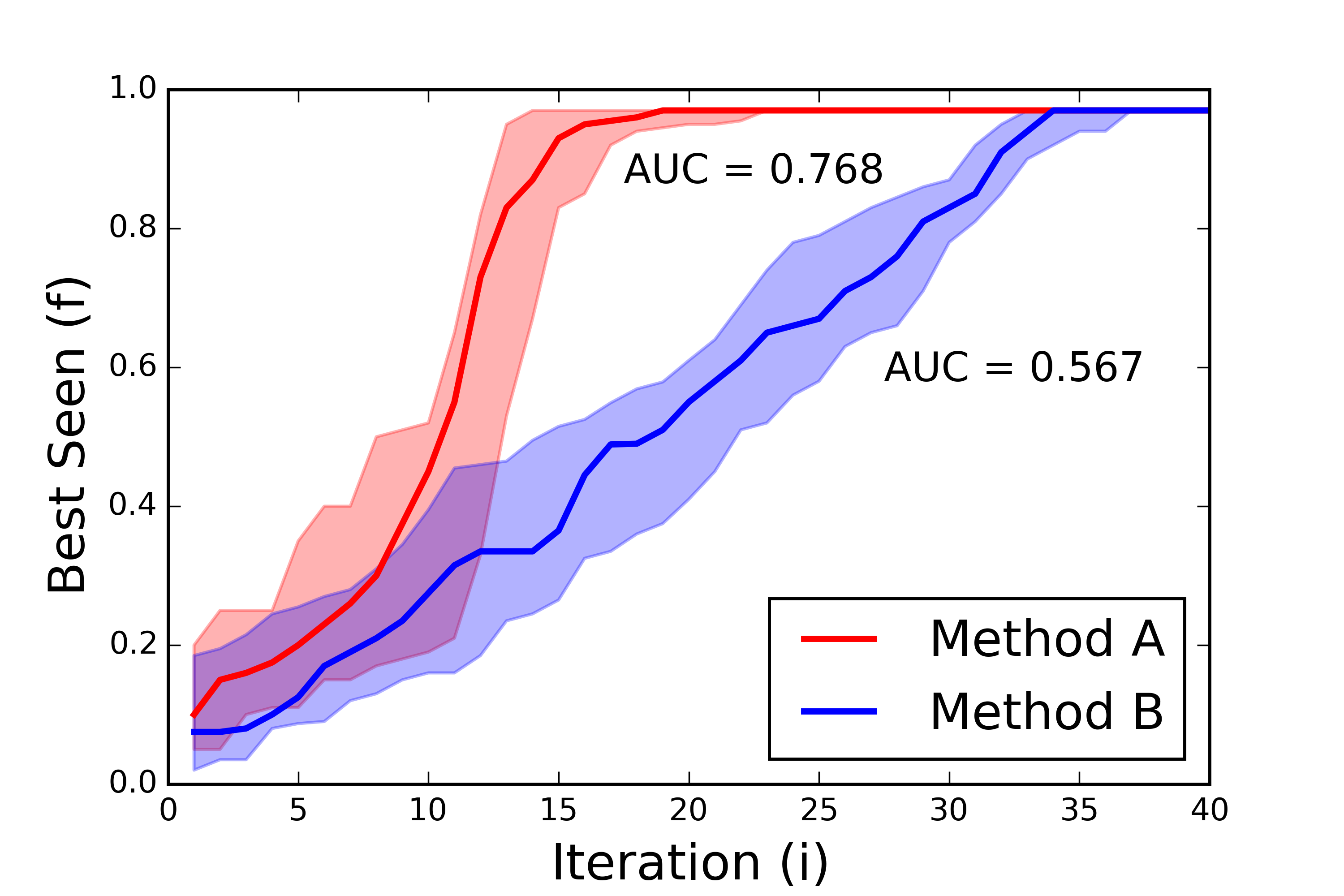}
	\caption{Hypothetical optimization methods A and B both achieve the same \textbf{Best Found} of 0.97 after 40 evaluations.  Method A however finds the optimum in fewer evaluations. \label{fig:test}}
\end{figure}

The SigOpt service aims to maximize objective functions.  The performance metrics we consider for comparisons on a given objective function are the best value seen by the end of the optimization ( \textbf{Best Found} ), and the area under the best seen curve ( \textbf{AUC} ).  The \textbf{AUC} metric can help to better differentiate performance, as shown in \figref{fig:test}.

The stochastic nature of the optimization algorithms under consideration require that the performance metrics be interpreted statistically.  That is, the optimization performance on a given function will inherently vary from one run to the next, so multiple runs on a given function are required to discern statistically significant changes.  Generally, optimization algorithms are run 20 times on each function and the distributions of the performance metrics are compared using the non-parametric Mann-Whitney $U$ test, which has been suggested in previous empirical studies of Bayesian optimization methods \cite{hutter2011sequential}.  Further discussion of these metrics and statistical analysis is presented in \cite{dewancker2016stratified}    

\section{Benchmark Suite}

\begin{figure}[ht]
	\centering
	%
	\includegraphics[width=\columnwidth]{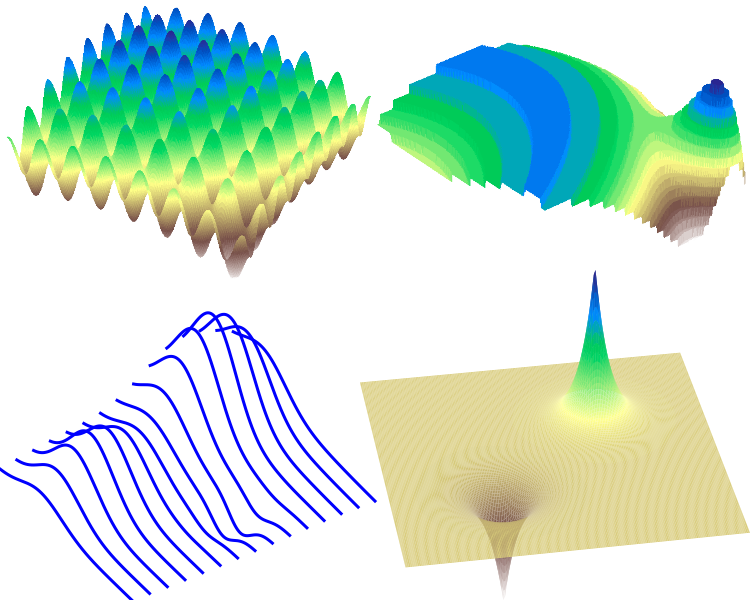}
	\caption{
		Sample benchmark functions. \textit{Top left}: Oscillatory,
		\textit{Top right}: Discrete valued,
		\textit{Bottom left}: Mixed integer,
		\textit{Bottom right}: Mostly boring,
		\label{fig:samplefunctions}
	}
\end{figure}

The tests for our evaluation system consist of closed-form optimization functions \cite{mccourt_test} which are extensions of an earlier set proposed by \cite{Gavana2013} for black-box optimization evaluation.  These functions are fast to evaluate and extensible.  We sought a collection that exhibited a wide variety of properties e.g. non-smooth, oscillatory.  Some representative functions and corresponding properties of interest are shown in Figure \ref{fig:samplefunctions}  

Design bias is an important concern when constructing any benchmark test suite or dataset.  One example of a design bias we initially encountered in our test suite was functions having optima in predictable locations, for example, at the domain midpoint or on integer coordinates.  In this benchmark suite we have made an effort to appropriately classify and segregate functions of this type, though further work is required to identify and resolve less obvious biases.

\section{Infrastructure}

Obtaining the empirical distributions of the performance metrics for every test function in our benchmark suite requires significant computational resources.  
Fortunately, these evaluation tasks are embarrassingly parallel since each function optimization can be done independently of
the others in the test suite, and each repeated run on the same function is also independent of other repeated runs.  

To co-ordinate this effort, lightweight function evaluation processes are run concurrently on a large master machine with many cores.  Each process communicates with an on-demand cluster of SigOpt API workers, which in turn co-ordinate each optimization request with a cluster of instances running the SigOpt optimization engine as well as a database used by the service.  The database persists important state for each optimization and is central to the production service.  Baseline optimization methods are run on the master machine directly. 

\begin{figure}[H]
	\centering
	\includegraphics[width=\columnwidth]{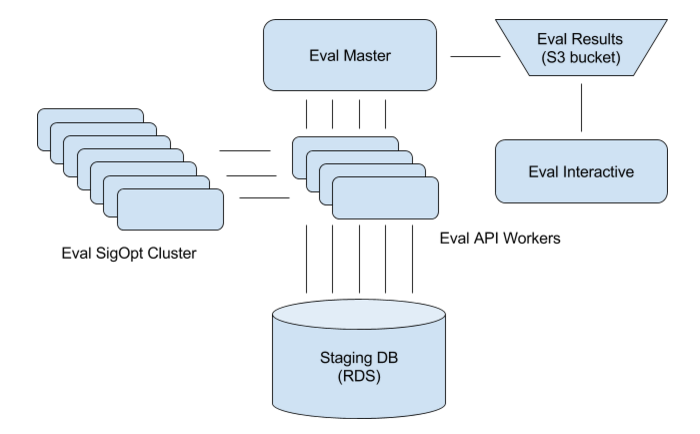}
	\caption{Architecture of evaluation system infrastructure  }
\end{figure}

Instances for the evaluation system are created as needed using cloud compute providers AWS and DigitalOcean.  
We found it was helpful to replicate our production optimization flow as closely as possible.  By re-creating much of the SigOpt production flow for the evaluation system, several issues and bugs were exposed relating to the API and database in addition to the core SigOpt optimization engine.  Results from every run are archived in a simple, extensible JSON format and stored in AWS S3.  An interactive web application is used to present evaluation summarizations and inspect results.  

\section{Visualization Tools}

Performance metrics and best seen traces are collected during each optimization run on all test functions.  In raw form, this information is daunting to summarize and extract actionable insights from.   
To assist in quickly summarizing these results and drilling down into the performance results on particular test functions we developed an interactive web application which hosts various visualizations of the evaluation data.

\subsection {Comparative Optimization Traces}

\begin{figure}[H]
	\label{fig:bestseentrace}
	\centering
	\includegraphics[width=\columnwidth]{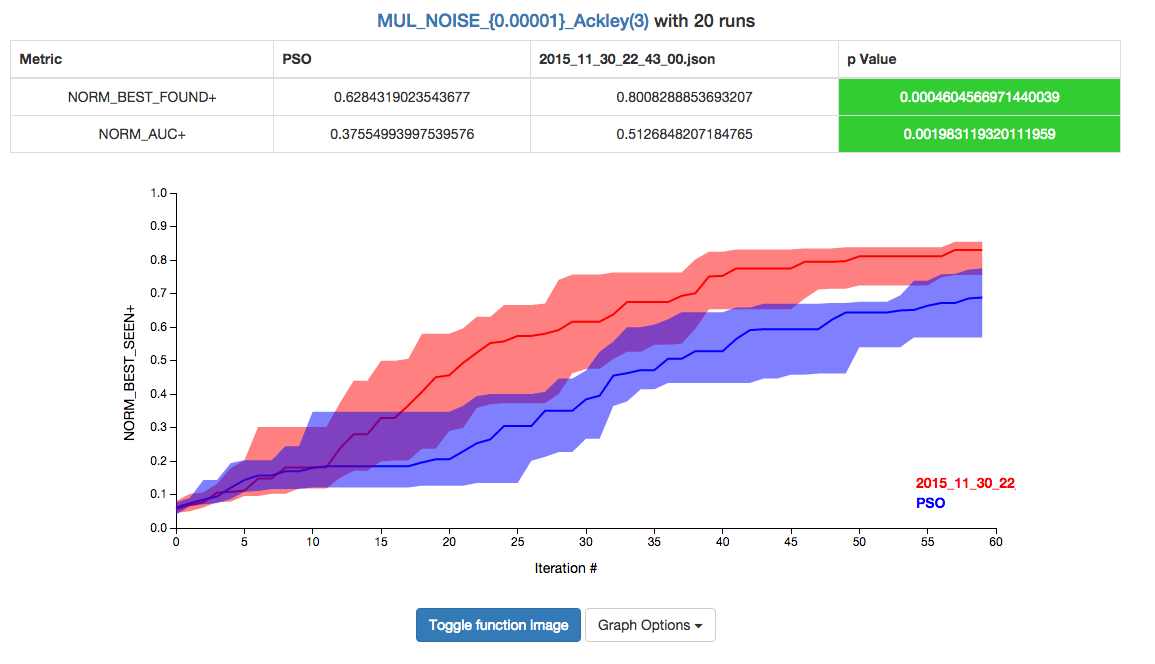}
	\caption{Visualization of best seen trace and metric summary for SigOpt and PSO on a given optimization test function }
\end{figure}

An important tool when diagnosing or comparing optimization efforts on a given function is the best seen trace.  The trace represents the best value of the objective metric seen after each function evaluation.  In the Bayesian optimization setting, each function evaluation is assumed to be expensive and so the efficiency of methods is most naturally compared using this measurement.  Each trace on a given function is stochastic, so the interquartile range of all traces and the median trace is plotted.
Traces are always produced in a comparative setting; either between two versions of SigOpt, or between SigOpt and an external optimization method. 
Figure \ref{fig:bestseentrace} shows a comparative version of SigOpt compared to a particle swarm optimization (PSO) implementation \cite{lee_pyswarm}.

A comparative metric summary table is also provided for each best seen trace visualization.  This table summarizes the results of the Mann-Whitney $U$ test performed on each method's metric distribution for the given test function.  Optimization method A is defined to be significantly better than method B on a given metric if the expected value of that metric is higher when using method A and the null hypothesis of the Man-Whitney $U$ test is rejected with statistical significance when comparing the two metric empirical distributions.  More formally, a win for method A over B using metric $M$ is defined as :
\begin{align*}
\text{\bf signf\_win}(A > B)_M =  \mathbb{E}[M_A] > \mathbb{E}[M_B]\   \land \\
                                                \text{\bf pval}(M_A, M_B) < 0.01
\end{align*}

We currently only consider the \textbf{Best Found} and \textbf{AUC} metrics, both of which are desired to be maximized, however our metric definition and reporting structure is extensible and can support metrics which are desired to be minimized.

\subsection{Comparative p-value Histograms}

\begin{figure}[H]
	\label{fig:pvalhist}
	\centering
	\includegraphics[width=\columnwidth]{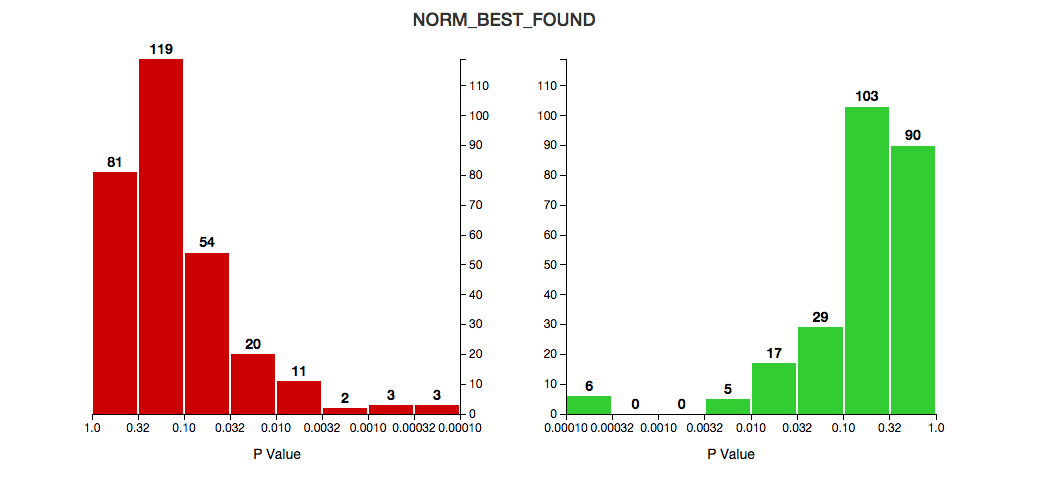}
	\includegraphics[width=0.9\columnwidth]{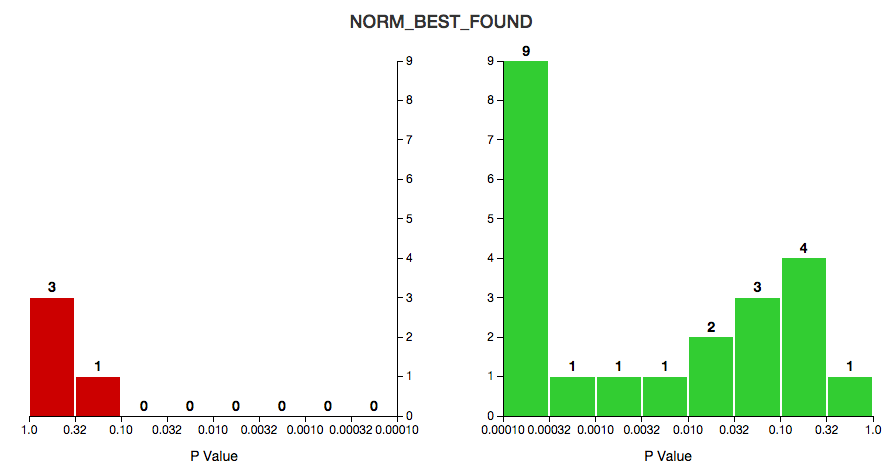}
	\caption{ \textit{Above}: Summary histograms for two optimization methods that were mostly comparable.  The test functions are evenly split and primarily binned to p-values for insignificant ranges. \textit{Below}: Summary histograms of two methods where method B (in green) shows a large number of test functions having p-values in the most significant bin, whereas method A (in red) shows only a few functions in the lowest significance ranges      }
\end{figure}

While the best seen traces are useful for inspecting performance on individual functions, it is also useful to have visualizations that help summarize the complete relative performances between two optimization methods on a given metric.  Towards this end, comparative histograms are generated representing the distribution of test functions over p-value ranges for a given metric.  For each metric $M$, we split the test functions into two sets and create two histograms of the p-values returned by the Mann-Whitney $U$ tests on the empirical distributions produced after evaluation runs.  Example histograms are shown in Figure \ref{fig:pvalhist}
\begin{align*}
\textbf{wins}( A > B )_{M} = \{ \ { \color{red} func } \ | \ \mathbb{E}[M_A] {\color{red} \ > \ } \mathbb{E}[M_B] \ \} \\
\textbf{wins}( B > A )_{M} = \{ \ {\color{ForestGreen} func } \ | \ \mathbb{E}[M_B] {\color{ForestGreen} \ > \ } \mathbb{E}[M_A] \ \}
\end{align*}

The histogram is interactive and each p-value bin can be clicked to inspect the best seen traces of all functions in that range.  This flow is particularly useful for investigating unexpected performance regressions or improvements on particular functions.  Intuitively, this chart visualizes a spread on the performance differences of two methods for a given metric.  A large number of functions binned in the center of the histogram implies many test functions exhibit significant metric performance differences.  Conversely, when more functions are allocated to the outer bins, this implies that most metric differences are not significant and the methods are probably comparable.

\subsection{Comparative Total Performance Tables }

The p-value histogram is useful for summarizing performance differences between methods on a given metric, however it is often useful to quickly understand a measure of the total relative performance between two methods summarized over all metrics.  Towards this end, we generate summary tables that count the number of wins, loses, ties and mixed performance comparisons between methods.  
An example table is show below in Table \ref{sample-table}

\begin{table}[th]
	\caption{Example total performance comparison of SigOpt to an external optimization method and a previous version of SigOpt}
	\label{sample-table}
	\vskip 0.15in
	\begin{center}
		\begin{small}
			\begin{sc}
				\begin{tabular}{ccccr}
					\hline
					\abovespace\belowspace
					SigOpt 2.01 (vs) &  Spearmint & SigOpt 1.85 \\
					\hline
					\abovespace
					Wins   & 65      &    8    \\
					Loses & 15      &    3      \\
					Ties    & 51     &  122  \\
					Mixed & 0        &   0   \\
					\hline
				\end{tabular}
			\end{sc}
		\end{small}
	\end{center}
	\vskip -0.1in
\end{table}

The total wins count represents the number of test functions where at least one metric has improved with statistical significance and all other metrics have not
changed with statistical significance.
\begin{multline*}
\text{\bf total\_wins}(A > B) = | \{ \ func \ | \ \exists M^{(1)} : \\ \mathbb{E}[M^{(1)}_A] > \mathbb{E}[M^{(1)}_B]\ \land                          
\text{\bf pval}(M^{(1)}_A, M^{(1)}_B) <= 0.01 \ \land  \\
( \ \forall M^{(2)} \ \neq M^{(1)}, \ \mathbb{E}[M^{(2)}_A] < \mathbb{E}[M^{(2)}_B] : \\ \text{\bf pval}(M^{(2)}_A, M^{(2)}_B) > 0.01 \ ) \} |
\end{multline*}

The total ties count is defined by the number of test functions where no metric has changed with statistical significance
\begin{multline*}
\text{\bf total\_ties}(A \,{==}\, B) = \\| \{ \ func \ | \  \forall M : \text{\bf pval}(M_A, M_B) > 0.01 \ \} |
\end{multline*}

Mixed results are functions where there exists one metric that increases with statistical significance and another that decreases with statistical significance.  
\begin{multline*}
\text{\bf total\_mixed}(A <> B) = | \{ \ func \ | \ \exists M^{(1)}, M^{(2)} : \\
 \mathbb{E}[M^{(1)}_A] > \mathbb{E}[M^{(1)}_B] \  \land \text{\bf pval}(M^{(1)}_A, M^{(1)}_B) <= 0.01 \ \land \\
\mathbb{E}[M^{(2)}_A] < \mathbb{E}[M^{(2)}_B] \ \land \text{\bf pval}(M^{(2)}_A, M^{(2)}_B) <= 0.01 \ \} |
\end{multline*}

\section{Conclusions}

Our evaluation system has been become a valuable analysis tool when considering algorithm or system changes to the SigOpt optimization service.  Data driven performance analysis is an effective way to enable faster iteration and evaluation of a wide spectrum of ideas.  The system continues to guide improvements to the core SigOpt service by providing empirical comparisons between internal changes and alternative methods from the Bayesian optimization community, as well helping to expose errors and bugs.  


\bibliography{example_paper}
\bibliographystyle{icml2016}

\end{document}